\documentclass{ieee}

\usepackage{times}
\usepackage{graphicx}
\usepackage{subfigure}

\begin{document}

\title{Spatiotemporal Gabor filters: a new method for dynamic texture recognition}

\author{Wesley Nunes Gon\c{c}alves, Bruno Brandoli Machado, Odemir Martinez Bruno \\
Physics Institute of S\~ao Carlos (IFSC) \\
University of S\~ao Paulo (USP)\\
Av. Trabalhador S\~ao-carlense, 400 \\ 
Cx. Postal 369 - S\~ao Carlos - SP - Brasil \\
wnunes@ursa.ifsc.usp.br, brandoli@icmc.usp.br, bruno@ifsc.usp.br \\
}

\maketitle
\thispagestyle{empty}

\begin{abstract}
This paper presents a new method for dynamic texture recognition based on spatiotemporal Gabor filters.
Dynamic textures have emerged as a new field of investigation that extends the concept of self-similarity of texture image to the spatiotemporal domain.
To model a dynamic texture, we convolve the sequence of images to a bank of spatiotemporal Gabor filters.
For each response, a feature vector is built by calculating the energy statistic.
As far as the authors know, this paper is the first to report an effective method for dynamic texture recognition using spatiotemporal Gabor filters.
We evaluate the proposed method on two challenging databases and the experimental results indicate that the proposed method is a robust approach for dynamic texture recognition.
\end{abstract}

%-------------------------------------------------------------------------
\section{Introduction}
The vision of animals provides a large amount of information that improves the perception of the world.
This information is processed into different dimensions, including color, shape, illumination, and motion.
While most of the features provide information about the static world, the motion provides essential information for interaction with external environment.
In recent decades, the perception and interpretation of motion have attracted a significant interest in computer vision community \cite{Qian2009,Wu2010,Cedras1995,Fazekas2007} motivated by the importance in both scientific and industrial communities.
Despite significant advances, the motion characterization is still an open problem.

For modeling image sequences, three classes of motion patterns have been suggested \cite{Polana1997}: dynamic texture, activities and events. 
The main difference between them relies on the temporal and spatial regularity of the motion field.
In this work, we aim at modeling dynamic textures, also called temporal textures.
They are basically texture in motion which is an extension of image texture to the spatiotemporal domain.
Examples of dynamic texture includes real world scenes of fire, flag blowing, sea waves, moving escalator, boiling water, grass, and steam.

Existing methods for dynamic texture can be classified into four categories according to how they model the sequence of images.
Due to the efficient estimation of features based on motion (e.g. optical flow), the motion based methods (i) are the most popular ones.
These methods model dynamic textures based on a sequence of motion patterns \cite{Fazekas2007,Polana1997,Fablet2003}.
For modeling dynamic texture at different scales in space and time, the spatiotemporal filtering based methods (ii) use spatiotemporal filters such as wavelet transform \cite{Dubois2009,Zhong2004,Dollar2005}.
Model based methods (iii) are generally based on linear dynamical systems, which provides a model that can be used in applications of segmentation, synthesis, and classification \cite{Doretto2003,Chan2007,Szummer1996}.
Based on properties of moving contour surfaces, spatiotemporal geometric property based methods (iv) extract motion and appearance features from the tangent plane distribution \cite{Fujii1998}.
The reader may consult \cite{Chetverikov2005} for a review of dynamic texture methods.
%Despite promising results achieved by recent methods, most of them present at least one of the following drawbacks: do not provide an explicit combination between motion features and appearance features, do not provide features that are robust against image transformations, cannot model multiple dynamic textures, and can be quite expensive.

In this paper, we propose a new approach for dynamic texture modeling based on spatiotemporal Gabor filters  \cite{Petkov2008}.
As far as the authors know, the present paper is the first one to model dynamic texture using spatiotemporal Gabor filters.
These filters are basically built using two parameters: the speed $v$ and direction $\theta$.
To model a dynamic texture, we convolve the sequence of images to a bank of spatiotemporal Gabor filter built with different values of speed and direction.
For each response, a feature vector is built by calculating the energy statistic.

We evaluate the proposed method by classifying dynamic texture from two challenging databases: dyntex \cite{dyntex} and traffic database \cite{Chan2005}.
Experimental results in both databases indicate that the proposed method is an effective approach for dynamic texture recognition.
For the dyntex database, filter with low speeds (e.g. 0.1 pixels/frame) achieved better results than high speeds.
In fact, the dynamic texture in this database presents low motion patterns.
On the other hand, for the traffic database, high speeds (e.g. 1.5 pixels/frame) achieved the best correct classification rate.
In this database, vehicles are moving at a speed that matches the filter's speed.

This paper is organized as follows.
Section 2 briefly describes spatiotemporal Gabor filters.
In Section 3, we present the proposed method for dynamic texture recognition based on spatiotemporal Gabor filters.
An analysis of the proposed method with respect to the speed and direction parameters is present in Section 4.
Experimental results are given in Section 5, which is followed by the conclusion of this work in Section 6.

\section{Spatiotemporal Gabor Filters}
Gabor filters are based on the important finding made by Hubel and Wiesel in the beginning of the 1960s. They found that the neurons of the primary visual cortex respond to lines or edges of a certain orientation in different positions of the visual field. Following this discovery, computational models were proposed for modeling the function of this neurons and the Gabor functions proved to be suited for this purpose in many works.

Initially, the researches aimed at studying spatial properties of the receptive field. However, some posterior studies revealed that cortical cells change in time and some of them are inseparable functions of space and time. Therefore, these cells are essentially spatiotemporal filters and they combine information over space and time, which makes a great model for dynamic texture analysis.

In this work, the spatiotemporal receptive field is modeled by a family of 3D Gabor filters \cite{Petkov2008} described in Equation \ref{eq:gabor}.

\setlength{\arraycolsep}{0.0em}
\begin{eqnarray}
\label{eq:gabor}
g_{(v,\theta,\varphi)}(x,y,t) &{}={}&\frac{\gamma}{2 \pi \sigma^2} \exp (\frac{-((\overline{x} + v_{c}t)^2 + \gamma^2 \overline{y}^2)}{2 \sigma^2}) \nonumber\\
&&.\cos(\frac{2 \pi}{\lambda} (\overline{x} + vt) + \varphi) \nonumber \\
&&.\frac{1}{\sqrt{2 \pi \tau}} \exp (\frac{-(t - \mu_t)^2}{2 \tau^2}) \\
&& \nonumber \\
&&\overline{x} = x \cos \theta + y \sin \theta \nonumber \\
&&\overline{y} = -x \sin \theta + y \cos \theta \nonumber 
\end{eqnarray}
\setlength{\arraycolsep}{5pt}

\begin{figure}[!t]
  \begin{center}
    \includegraphics[width=0.32\columnwidth]{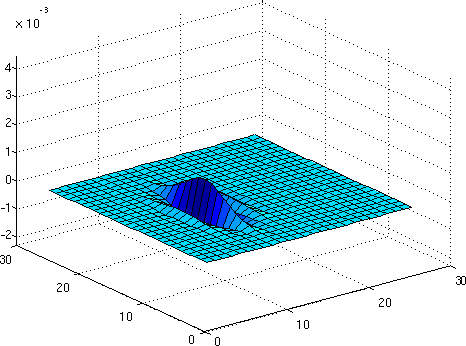}
    \includegraphics[width=0.32\columnwidth]{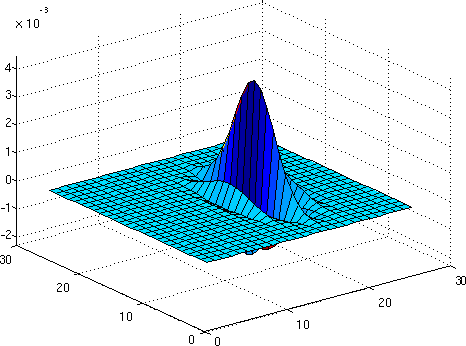}
    \includegraphics[width=0.32\columnwidth]{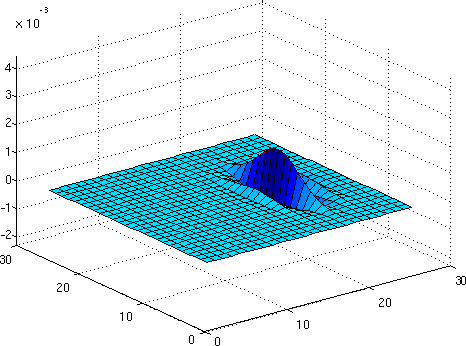} \\
    \includegraphics[width=0.32\columnwidth]{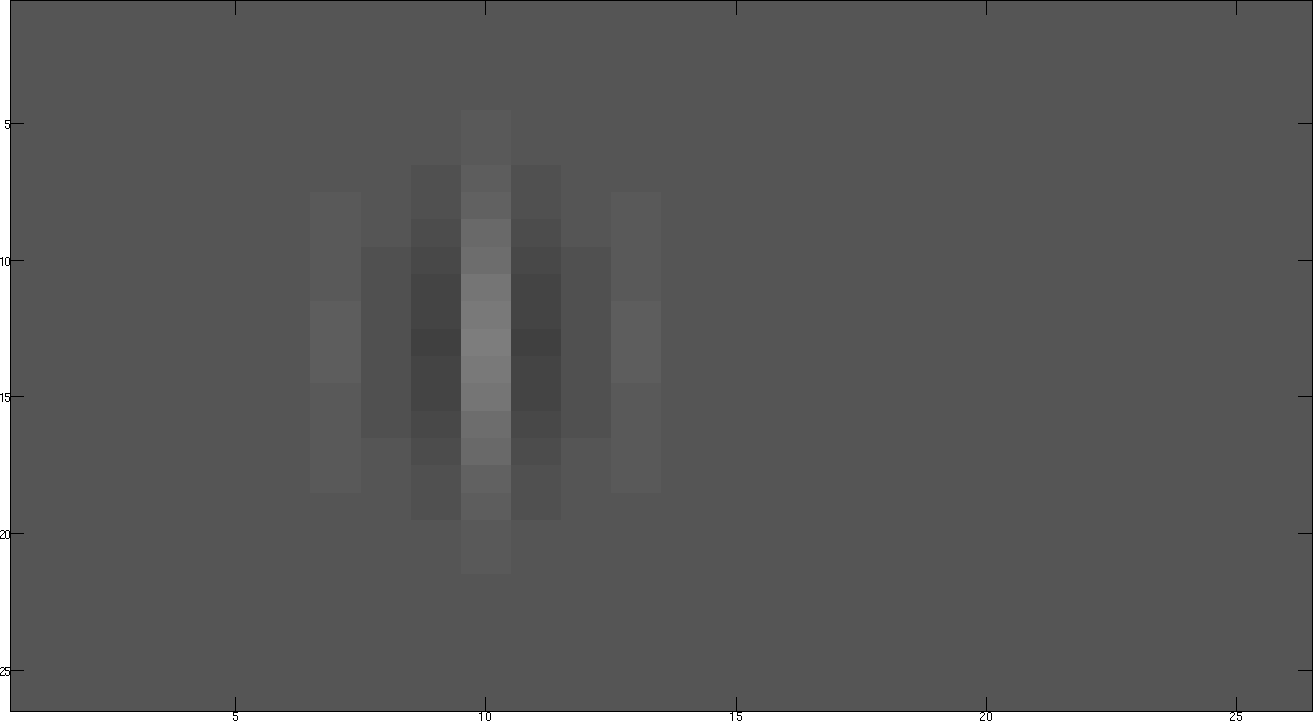}
    \includegraphics[width=0.32\columnwidth]{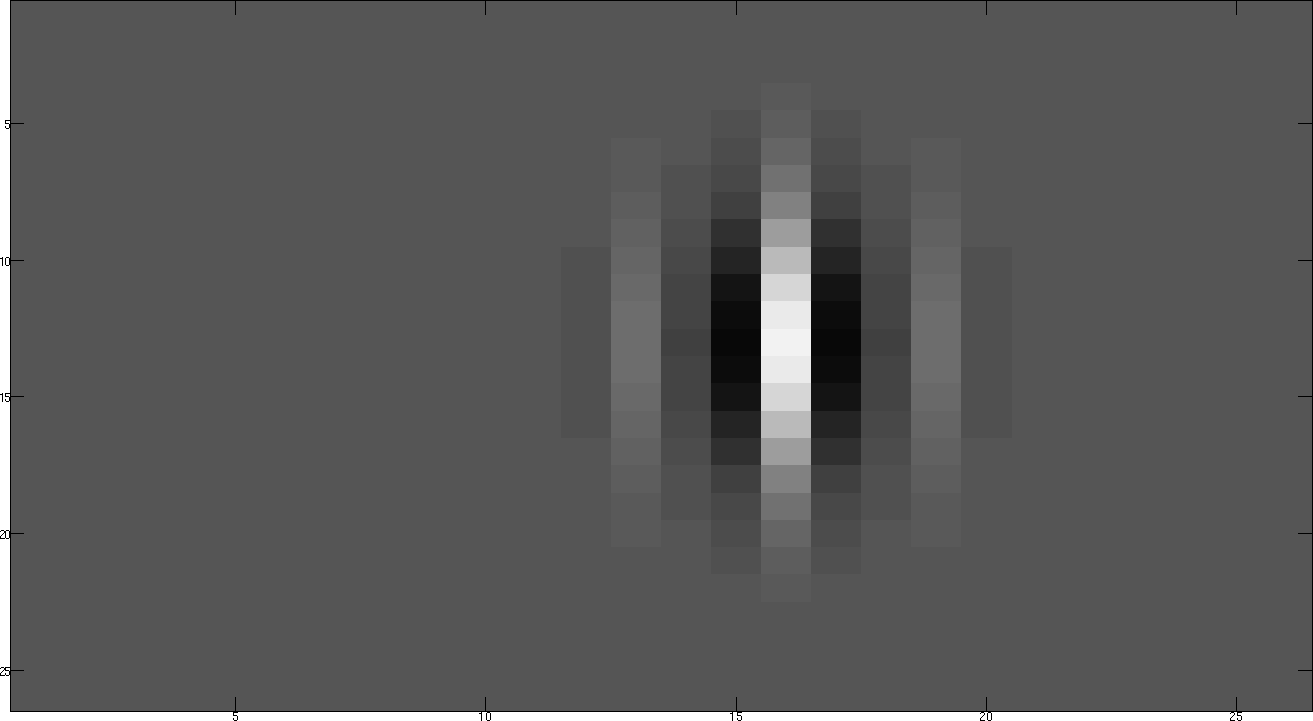}
    \includegraphics[width=0.32\columnwidth]{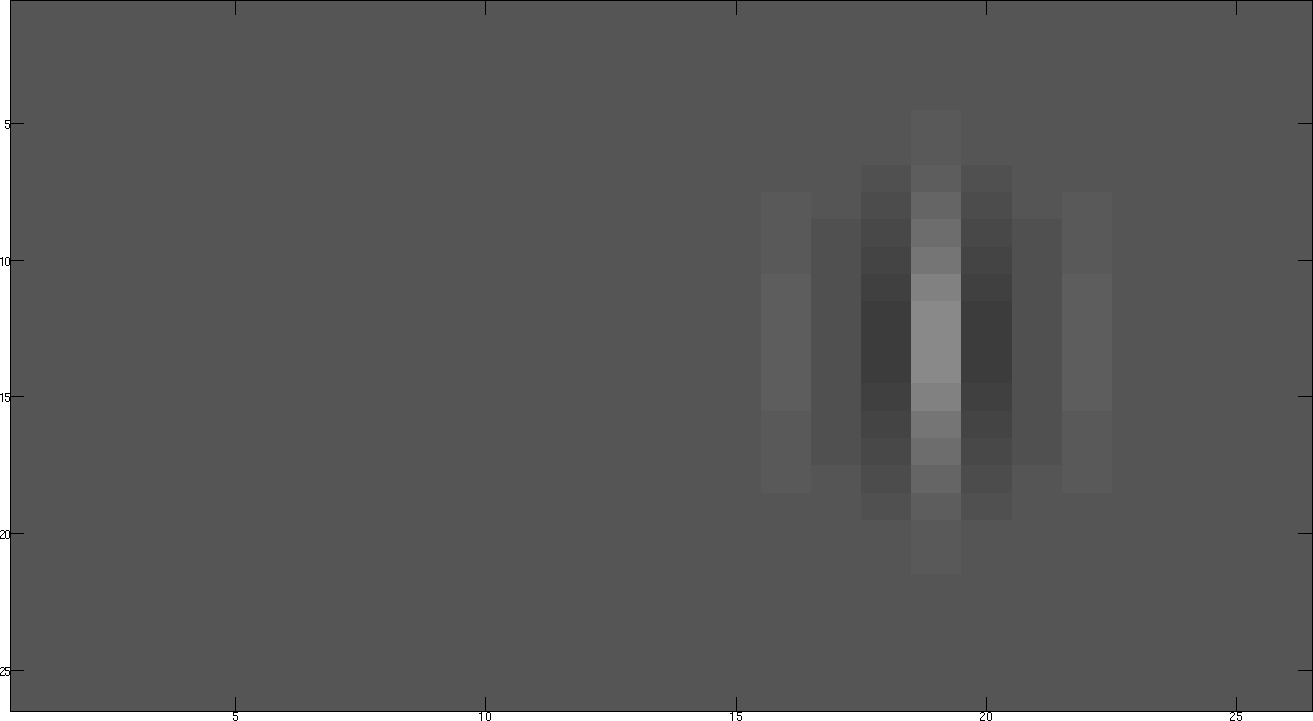} \\
    \caption{Example of spatiotemporal Gabor filter for $v_c = 1$.}
    \label{fig:movingEnvelope}
  \end{center}
\end{figure}

We now discuss the parameters of the spatiotemporal Gabor filter.
Some parameters were empirically found based on studies of response in the receptive visual field \cite{Petkov2008}.
The size of the receptive field is determined by the standard deviation $\sigma$ of the Gaussian factor.
The parameter $\gamma$ is the rate that specifies the ellipticity of the Gaussian envelope in the spatial domain.
This parameter is set to $\gamma = 0.5$ for match to the elongated receptive field along $\overline{y}$ axis.
The speed $v$ is the phase speed of the cosine factor, which determines the speed of motion.
The speed which the center of the spatial Gaussian moves along the $\overline{x}$ axis is specified by the parameter $v_c$.
When $v_c = 0$, the center of the Gaussian envelope is stationary.
On the other hand, a moving envelope is obtained when $v_c = v$.
Figure \ref{fig:movingEnvelope} presents a moving envelope with $v_c = v = 1$.

The parameter $\lambda$ is the wavelength of the cosine factor.
It is obtained through the relation $\lambda = \lambda_0 \sqrt{1 + v^2}$, where the constant $\lambda_0 = 2$.
The angle $\theta \in [0, 2 \pi]$ determines the direction of motion and the spatial orientation of the filter.
The phase offset $\varphi \in [-\pi, \pi]$ determines the symmetry in the spatial domain.
The Gaussian distribution with mean $\mu_t$ and standard deviation $\tau$ is used to model the change in intensities.
The mean $\mu_t = 1.75$ and standard deviation $\tau = 2.75$, both parameters are fixed based on the mean duration of most receptive fields.

\section{Dynamic Textures Modeling based on Spatiotemporal Gabor Filters}
In this section, we describe the proposed method for dynamic texture modeling based on spatiotemporal Gabor filters.
Briefly, the sequence of images is convolved with a bank of spatiotemporal Gabor filters and a feature vector is constructed with the energy of the responses as components.

The response $r_{(v,\theta,\varphi)}(x,y,t)$ of a spatiotemporal Gabor filter $g_{(v,\theta,\varphi)}$ to a sequence of images $I(x,y,t)$ is computed by convolution:

\begin{equation}
r_{(v,\theta,\varphi)}(x,y,t) = I(x,y,t) * g_{(v,\theta,\varphi)}
\end{equation}

Spatiotemporal Gabor filters are phase sensitive because its response to a moving pattern depends on the exact position within the sequence of images.
To overcome this drawback, a response that is phase insensitive can be obtained by:
\begin{equation}
R_{(v,\theta)} = \sqrt{r^{2}_{(v,\theta,\varphi)}(x,y,t) + r^{2}_{(v,\theta,\varphi/2)}(x,y,t)}
\end{equation}

To characterize the Gabor space resulting from the convolution, the energy of the response is computed according to Equation \ref{eq:energy}.
\begin{equation}
\label{eq:energy}
E_{(v,\theta)} = \sum_{x,y,t} R_{(v,\theta)}^{2}(x,y,t)
\end{equation}

A central issue in applying spatiotemporal Gabor filters is the determination of the filter parameters that covers the spatiotemporal frequency space, and captures dynamic texture information as much as possible.
Each spatiotemporal Gabor filter is determined by two main parameters: the direction $\theta$ and the speed of motion $v$.
In order to cover a wide range of dynamic textures, we design a bank of spatiotemporal Gabor filter using a set of values for speed $V = [v_1, v_2, \dots, v_n]$ and direction $\Theta = [\theta_1, \theta_2, \dots, \theta_m]$.
The feature vector that characterizes the dynamic texture is composed by the energy of the response for each combination of velocities $V$ and direction $\Theta$ (Equation \ref{eq:featureVector}).

\begin{equation}
\label{eq:featureVector}
\varphi = [E_{(v_1,\theta_1)}, \dots, E_{(v_n,\theta_m)}]
\end{equation}

\begin{figure}[!t]
  \begin{center}
    \includegraphics[width=1\columnwidth]{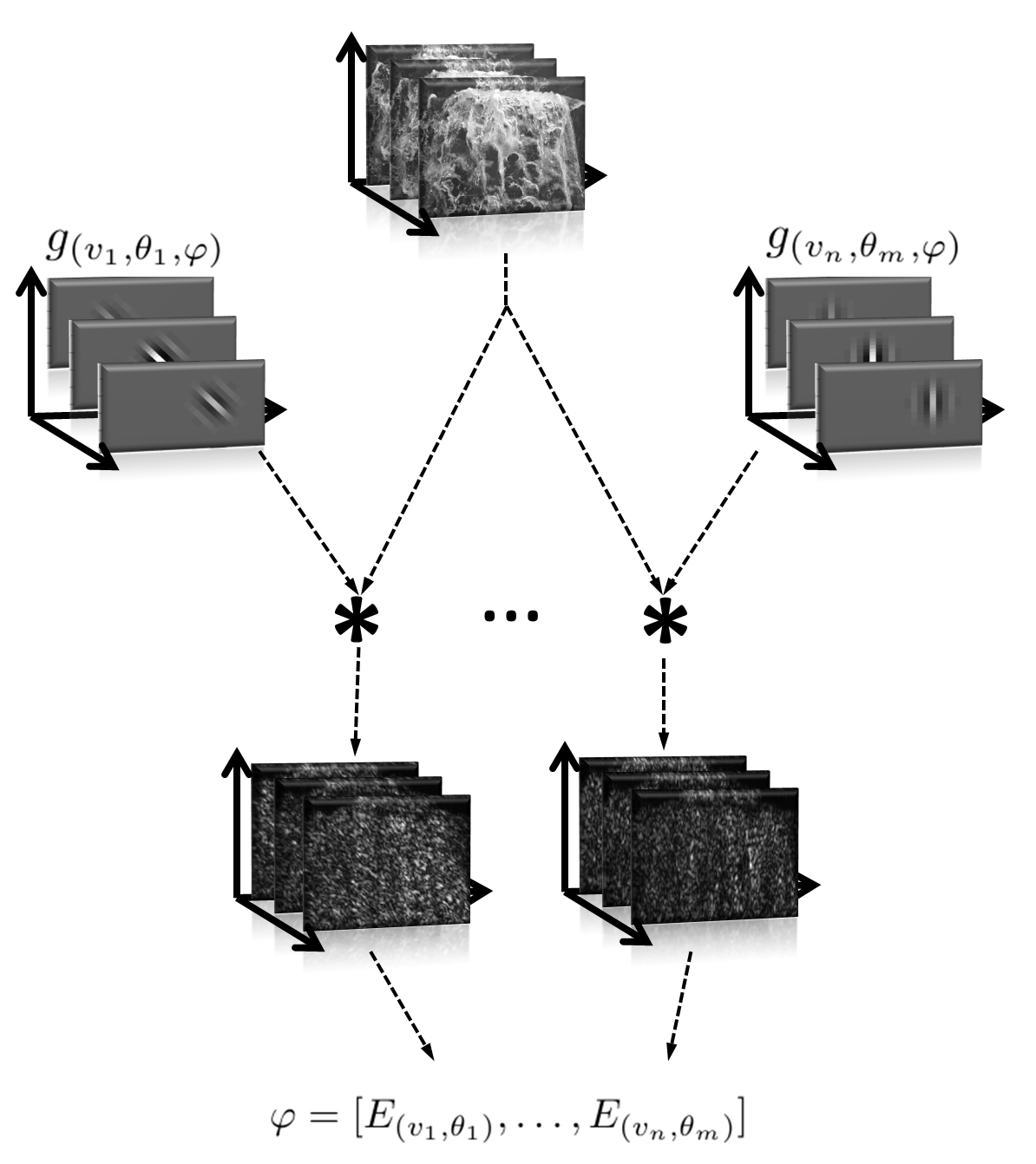}
    \caption{Proposed method considers the following steps:
(i) Design a bank of spatiotemporal Gabor filters using different values of speed $v$ and direction $\theta$; 
(ii) Convolve the sequence of images with the bank of filters; 
(iii) Calculate the energy for each response to compose the feature vector.}
    \label{fig:metodo}
  \end{center}
\end{figure}

The proposed method is summarized in Figure \ref{fig:metodo}.
First, we design a bank of spatiotemporal Gabor filters composed by filters with different directions and speeds.
Then, the sequence of images is convolved with the bank of filters.
For each convolved sequence of images, we calculate the energy to compose a feature vector.

\section{Response Analysis of Spatiotemporal Gabor Filters}
\begin{figure}[!t]
  \begin{center}
    \includegraphics[width=0.22\columnwidth]{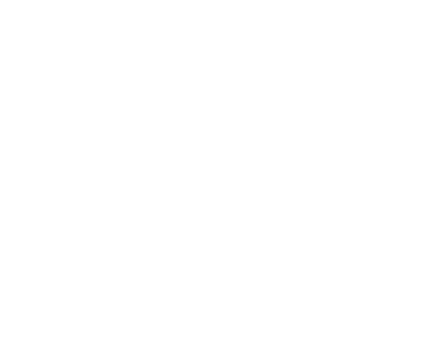}
    \includegraphics[width=0.22\columnwidth]{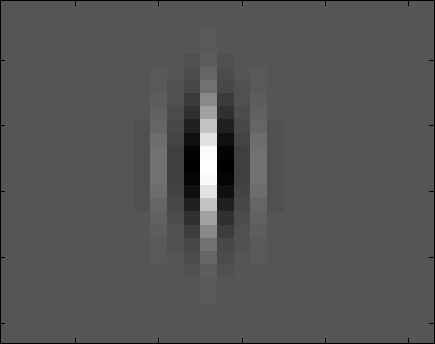}
    \includegraphics[width=0.22\columnwidth]{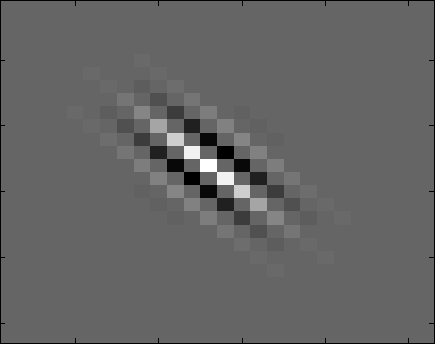}
    \includegraphics[width=0.22\columnwidth]{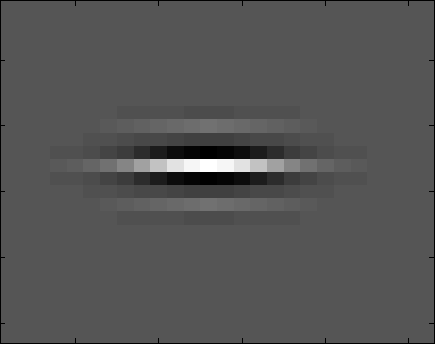}\\
    \includegraphics[width=0.22\columnwidth]{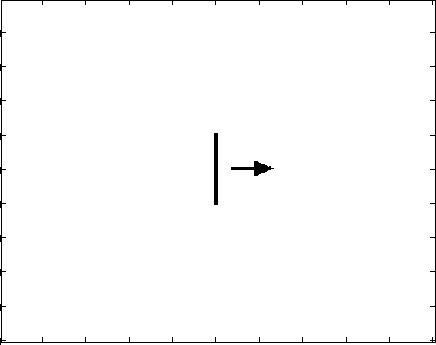}
    \includegraphics[width=0.22\columnwidth]{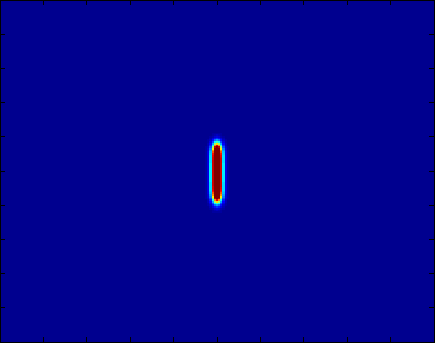}
    \includegraphics[width=0.22\columnwidth]{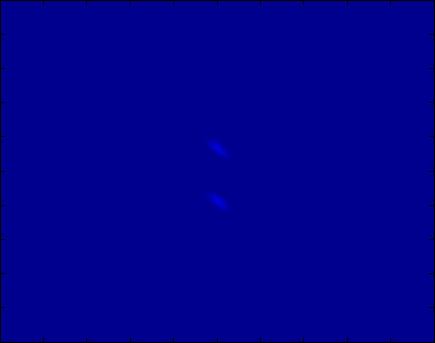}
    \includegraphics[width=0.22\columnwidth]{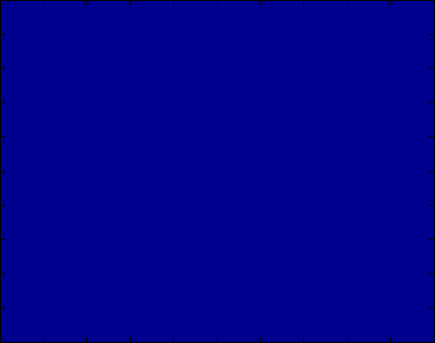}\\
    \includegraphics[width=0.22\columnwidth]{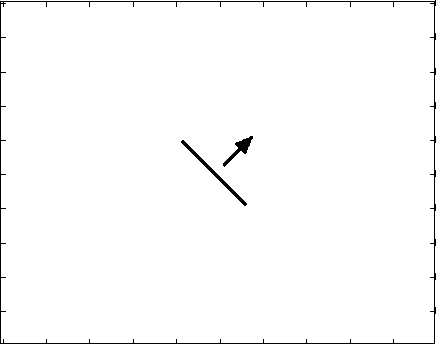}
    \includegraphics[width=0.22\columnwidth]{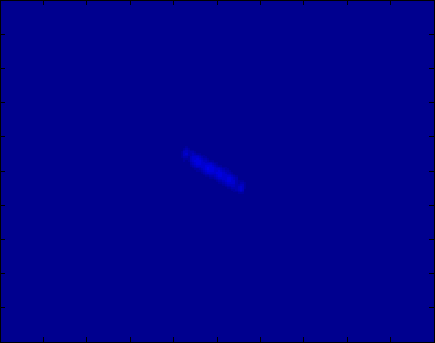}
    \includegraphics[width=0.22\columnwidth]{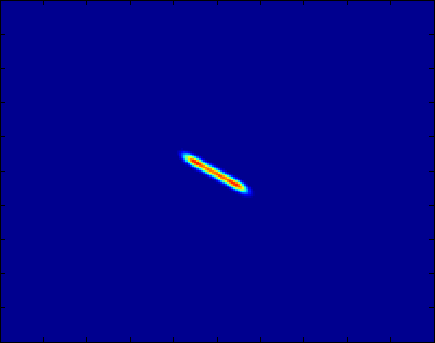}
    \includegraphics[width=0.22\columnwidth]{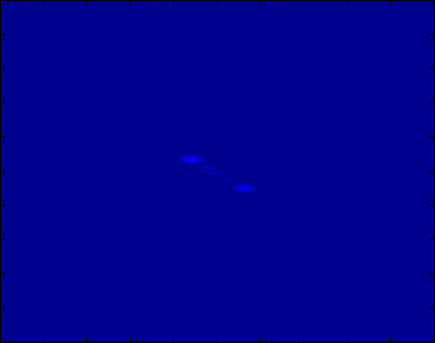}\\
    \includegraphics[width=0.22\columnwidth]{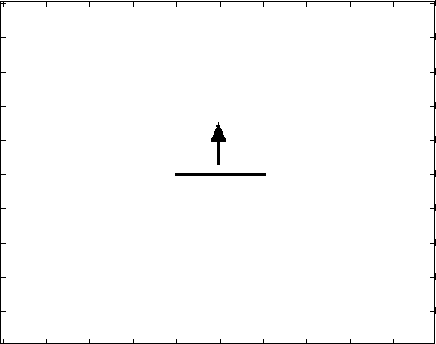}
    \includegraphics[width=0.22\columnwidth]{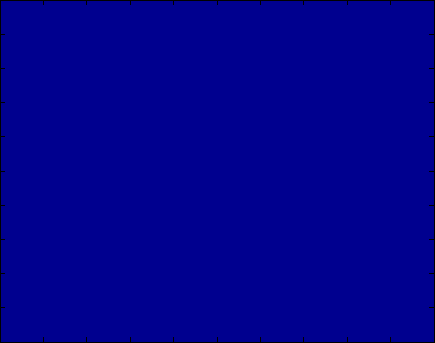}
    \includegraphics[width=0.22\columnwidth]{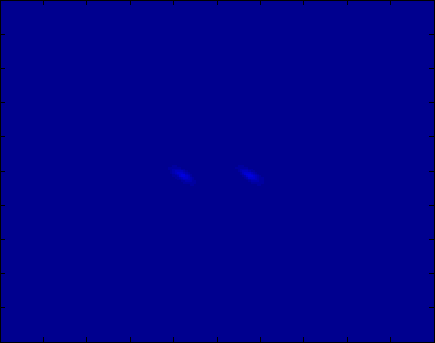}
    \includegraphics[width=0.22\columnwidth]{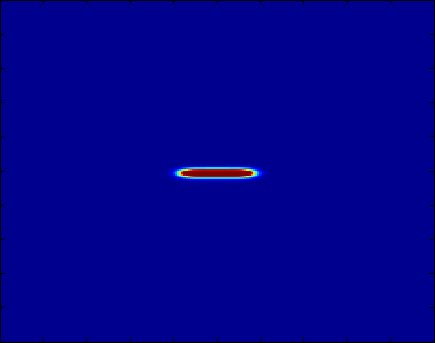} \\
    \caption{Response of spatiotemporal Gabor filters to bars moving in different direction $\theta$. First row corresponds to the filters. Second, third and fourth rows corresponds to the response of a bar moving in direction $\theta = 0, \frac{\pi}{4}, \frac{\pi}{2}$, respectively.}
    \label{fig:directionResponse}
  \end{center}
\end{figure}

Here, we analyze the speed and direction properties of the spatiotemporal Gabor filters in synthetic sequence of images. In Figure \ref{fig:directionResponse}, we present the response of spatiotemporal Gabor filters to bars moving at the same speed but in different direction $\theta$.
The filters and the moving bars have preference for the same speed $v = 1$.
The response has the highest magnitude when the direction of the filter $\theta$ matches the direction of the moving bar.
For instance, when $\theta = 0$, a vertical bar moving rightwards evokes higher response than bars with other direction of movement.

\begin{figure}[!t]
  \begin{center}
    \includegraphics[width=0.22\columnwidth]{white.png} 
    \includegraphics[width=0.22\columnwidth]{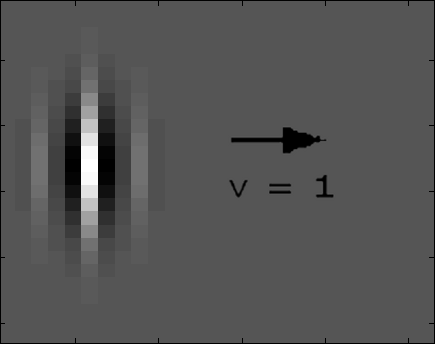}
    \includegraphics[width=0.22\columnwidth]{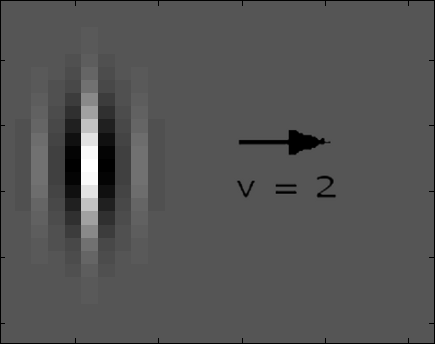}
    \includegraphics[width=0.22\columnwidth]{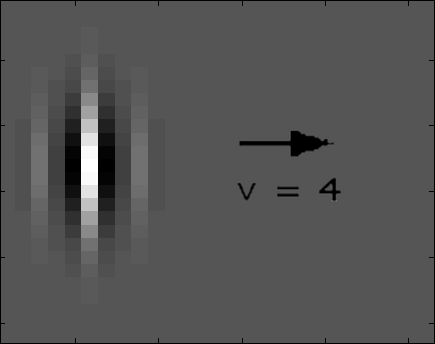}\\
    \includegraphics[width=0.22\columnwidth]{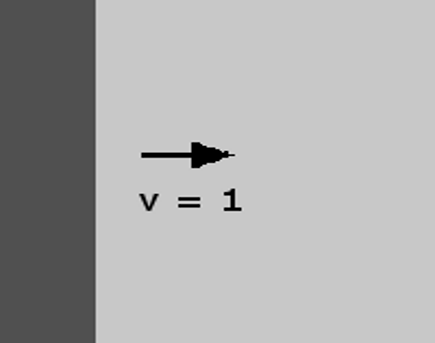}
    \includegraphics[width=0.22\columnwidth]{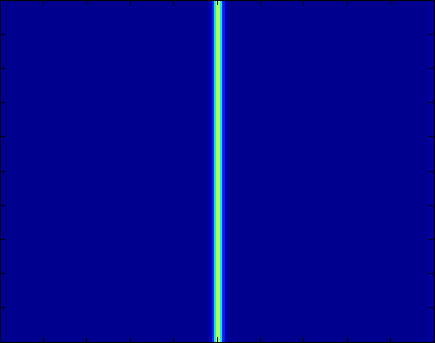}
    \includegraphics[width=0.22\columnwidth]{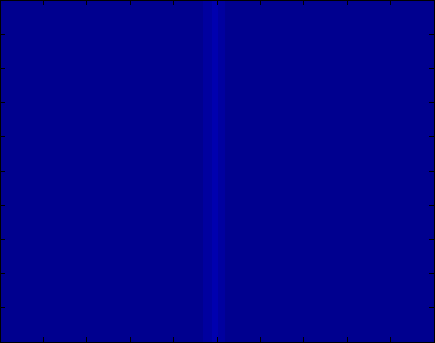}
    \includegraphics[width=0.22\columnwidth]{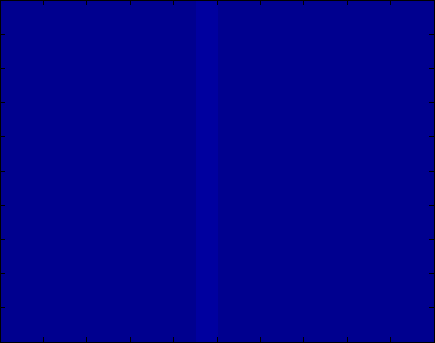}\\
    \includegraphics[width=0.22\columnwidth]{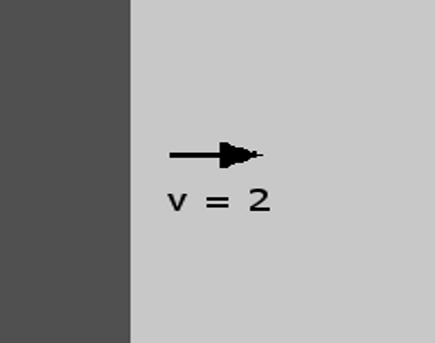}
    \includegraphics[width=0.22\columnwidth]{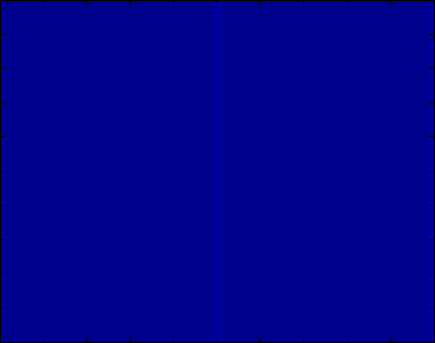}
    \includegraphics[width=0.22\columnwidth]{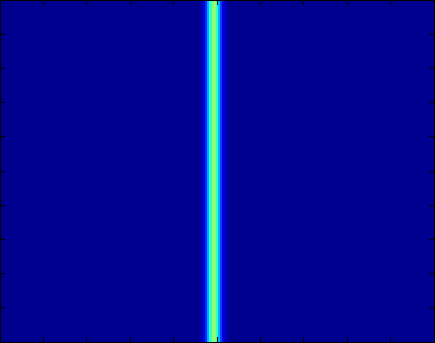}
    \includegraphics[width=0.22\columnwidth]{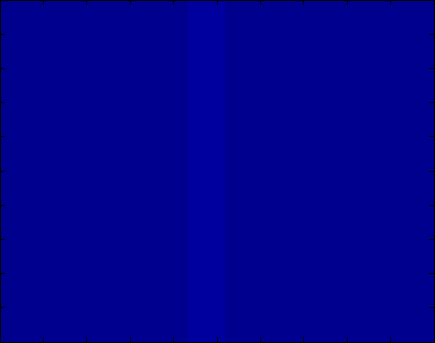}\\
	\includegraphics[width=0.22\columnwidth]{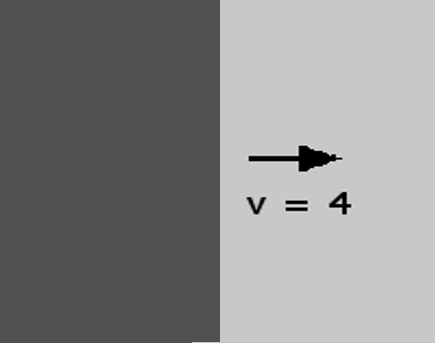}
    \includegraphics[width=0.22\columnwidth]{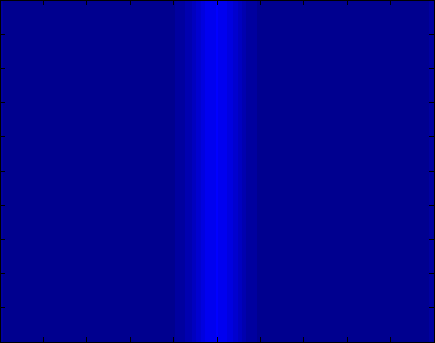}
    \includegraphics[width=0.22\columnwidth]{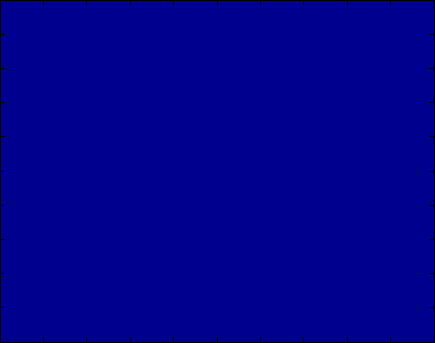}
    \includegraphics[width=0.22\columnwidth]{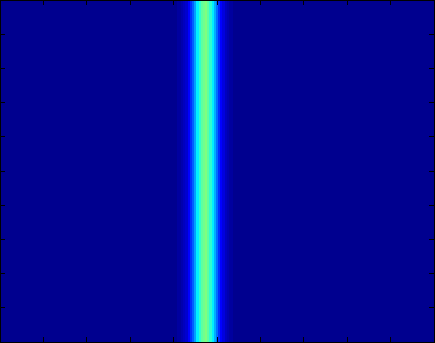}\\
    \caption{Response of spatiotemporal Gabor filters to edges moving at different speed $v$. First, second, and third rows corresponds to the response of a edges moving at speed $v = 1, 2, 4$, respectively.}
    \label{fig:speedResponse}
  \end{center}
\end{figure}

The speed property is evaluated in Figure \ref{fig:speedResponse}.
We analyze the response of spatiotemporal Gabor filters to edges drifting rightward at different speeds.
The filters and the synthetic sequence of images have for the same preference direction $\theta = 0$.
The highest response is achieved by filters which the speed matches the speed of the moving edge.

%\begin{figure}[!t]
%  \begin{center}
%    \includegraphics[width=0.32\columnwidth]{figures/vidf3_33_028_f030.png} 
%%    \includegraphics[width=0.24\columnwidth]{figures/vidf3_33_028_f040.png}
%    \includegraphics[width=0.32\columnwidth]{figures/vidf3_33_028_f050.png}
%    \includegraphics[width=0.32\columnwidth]{figures/vidf3_33_028_f060.png}\\
%    \includegraphics[width=0.32\columnwidth]{figures/response_0_f=30.png}
% %   \includegraphics[width=0.24\columnwidth]{figures/response_0_f=40.png}
%    \includegraphics[width=0.32\columnwidth]{figures/response_0_f=50.png}
%    \includegraphics[width=0.32\columnwidth]{figures/response_0_f=60.png}\\
%    \includegraphics[width=0.32\columnwidth]{figures/response_pi_4_f=30.png}
%  %  \includegraphics[width=0.24\columnwidth]{figures/response_pi_4_f=40.png}
%    \includegraphics[width=0.32\columnwidth]{figures/response_pi_4_f=50.png}
%    \includegraphics[width=0.32\columnwidth]{figures/response_pi_4_f=60.png}\\
%	\includegraphics[width=0.32\columnwidth]{figures/response_pi_f=30.png}
%   % \includegraphics[width=0.24\columnwidth]{figures/response_pi_f=40.png}
%    \includegraphics[width=0.32\columnwidth]{figures/response_pi_f=50.png}
%    \includegraphics[width=0.32\columnwidth]{figures/response_pi_f=60.png}\\
%    \caption{XX.}
%    \label{fig:speedResponse}
%  \end{center}
%\end{figure}

\begin{figure*}[!t]
  \begin{center}
    \includegraphics[width=0.48\textwidth]{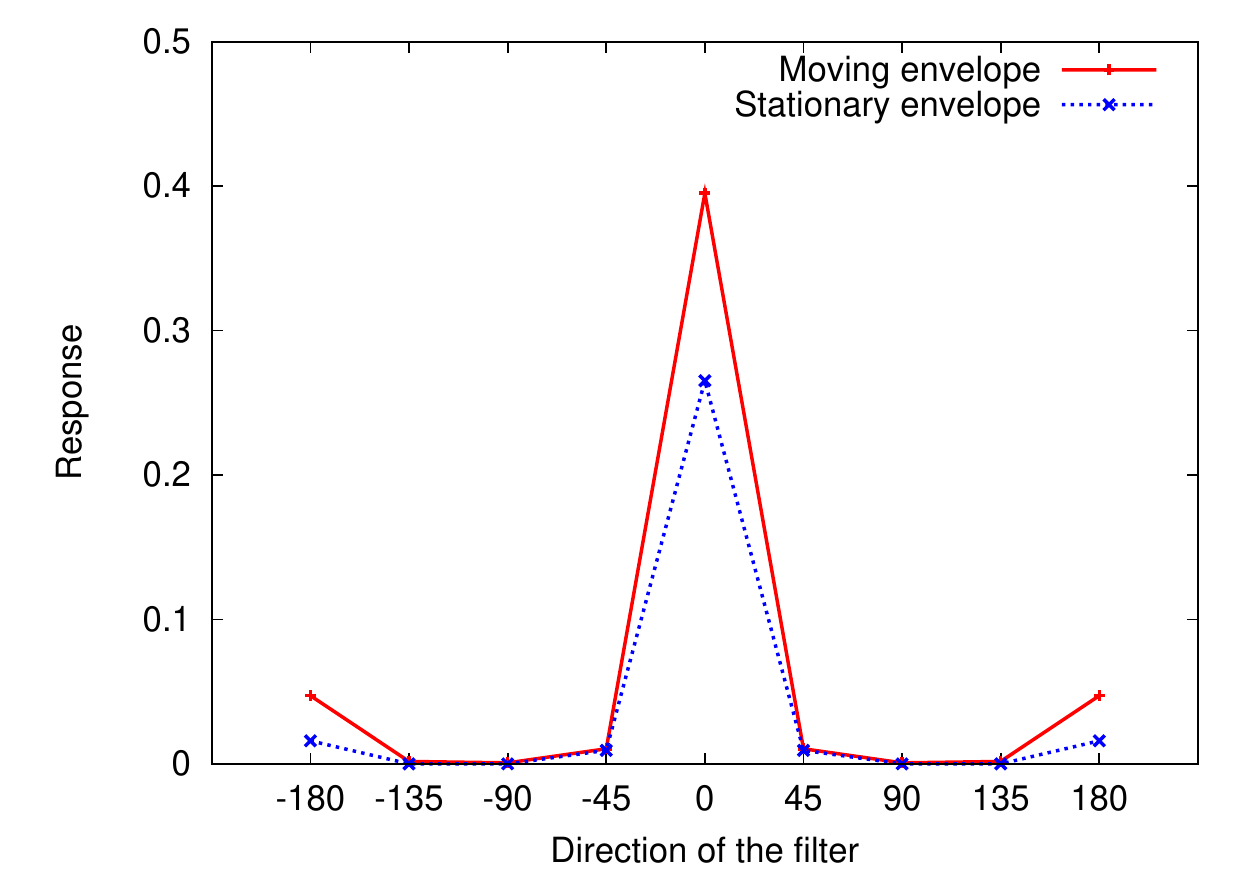}
	\includegraphics[width=0.48\textwidth]{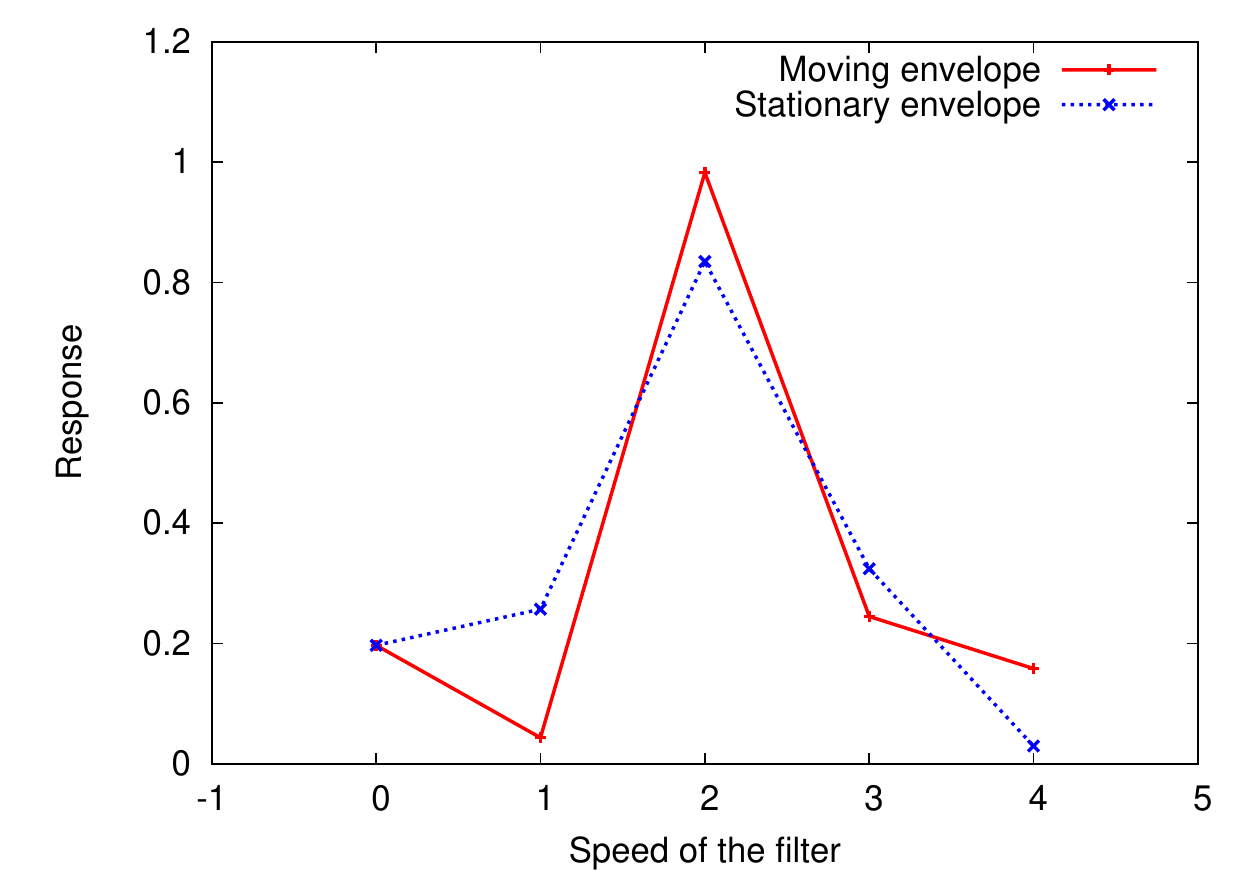}
    \caption{(a) Response for filters of different direction $\theta$ to a moving bar with $\theta = 0$.
    (b) Response for filters of different speeds $v$ to a edge moving at speed $v=2$.}
    \label{fig:speed_direction}
  \end{center}
\end{figure*}

In Figure \ref{fig:speed_direction}(a), we plot the response of filters to a moving bar with direction $\theta=0$ at a speed $v=1$.
The response reaches its maximum value to a filter with direction $\theta = 0$, which matches to the direction of the moving bar.
As we can see, the filter with moving envelope ($v_c = v$) achieved higher response than a filter with stationary envelope ($v_c = 0$).
The plot for the speed parameter is shown in Figure \ref{fig:speed_direction}(b).
We convolved filters to an edge drifting rightward in direction $\theta = 0$ at a speed of $v=2$.
The maximum response is achieved for the filter whose speed matches to stimulus' speed.
Again, we can conclude that filters with the moving envelope are more selective for both direction and speed than filters with stationary envelope.

\section{Experimental Results}
\begin{figure}[!t]
   \begin{center}
     \includegraphics[width=0.9\columnwidth]{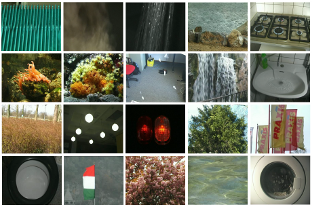}
     \caption{Examples of dynamic textures from dyntex database \cite{dyntex}. The database is composed by 50 dynamic texture classes each containing 10 samples.}
     \label{fig:dyntexdatabase}
   \end{center}
\end{figure}

\begin{figure}[!t]
   \begin{center}
     \includegraphics[width=0.8\columnwidth]{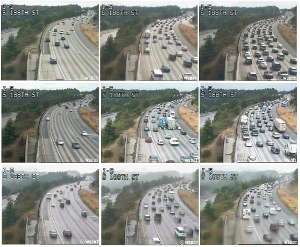}
     \caption{Examples of dynamic textures from traffic database \cite{Chan2005}. The database consists of 254 videos split into three classes of light, medium and heavy traffic.}
     \label{fig:trafficdatabase}
   \end{center}
\end{figure}

In this section, we present the experimental results using two databases: (i) dyntex database and (ii) traffic video database.
The dyntex database consists of 50 dynamic texture classes each containing 10 samples collected from Dyntex database \cite{dyntex}.
The videos are at least 250 frames long with dimension of $400 \times 300$ pixels.
Figure \ref{fig:dyntexdatabase} shows examples of dynamic textures from the first database.
The second database, collected from traffic database \cite{Chan2005}, consists of 254 videos divided into three classes: light, medium, and heavy traffic.
Videos had 42 to 52 frames with a resolution of $320 \times 240$ pixels.
The variety of traffic patterns and weather conditions are shown in Figure \ref{fig:trafficdatabase}.
All the experiments used a k-nearest neighbor classifier with $k=1$ in a scheme 10-fold cross validation.

Now, we discuss the influence of the direction and speed parameters in the dynamic texture recognition.
Table \ref{tab:traffic} shows the average and standard deviation of correct classification rate for the traffic database.
Columns present the direction parameter evaluation using 4-directions, which is a combination of filters with $\theta = [0,  \frac{\pi}{2}, \pi, \frac{3\pi}{2}]$, and 8-directions which is a combination of filters with $\theta = [0, \frac{\pi}{4}, \frac{\pi}{2}, \frac{3\pi}{4}, \pi, \frac{5\pi}{4}, \frac{3\pi}{2}, \frac{7\pi}{4}]$.
Rows present the speed evaluation using combination of speed with step of 0.25.
As we can see, the bank of filter composed by filters with 8-directions outperformed the bank of filter with 4-directions for all combination of speeds.
However, very little improvement can be appreciated as the combination of direction is increased.
The improvement in correct classification rate was on average less than $1\%$ when direction combination rises from 4 to 8.
This is because the cars on the highway move always on the same direction, which can be modeled by 4-direction.
With respect to the speed parameter, the best results were achieved for high speeds, such as 1.5 pixels/frame and 2.0 pixel/frame.
A correct classification rate of $91.50\%$ was achieved by a bank of filter composed by 8-directions and speeds $[0.5, 0.75, 1.0, 1.25, 1.5, 1.75, 2.0]$.
The high speeds match to speed of the cars in the sequence of images and then the traffic jam can be modeled using these parameters.

\begin{table}[!t]
	\centering
		\begin{tabular}{|c|c|c|c|c|}
%			\cline{1-7} \cline{10-16}
			\hline
			\hline
			Speed Combination & 4-directions & 8-directions \\
			\hline
			$[0.50 - 0.75]$ & 90.06(5.53) & 90.18(5.62) \\
			$[0.50 - 1.00]$ & 90.17(5.70) & 90.57(5.15) \\ 
			$[0.50 - 1.25]$ & 89.07(5.91) & 89.90(5.06) \\
			$[0.50 - 1.50]$ & 89.68(5.99) & 89.74(5.80) \\
			$[0.50 - 1.75]$ & 89.69(6.18) & 90.56(5.58) \\
			$[0.50 - 2.00]$ & 89.45(5.77) & \textbf{91.50(5.20)} \\
			$[0.50 - 2.25]$ & 90.24(5.51) & 90.87(5.07) \\
			$[0.50 - 2.50]$ & 89.33(5.60) & 90.75(5.26) \\
			$[0.50 - 2.75]$ & 89.76(5.44) & 90.35(5.44) \\
			$[0.50 - 3.00]$ & 89.60(5.16) & 90.63(5.21) \\
			\hline
			%\cline{1-7} \cline{10-16}			
		\end{tabular}
	\caption{Correct classification rate and standard deviation for different combinations of speed and direction on the traffic database.}
	\label{tab:traffic}
\end{table}

In Table \ref{tab:dyntex}, we present the experimental results obtained on the dyntex database.
The same combination of directions of the early experiment was used to evaluate the proposed method.
However, as the dynamic textures in this database present low speeds, the combination of speed started in $v = 0.1$ pixels/frame and taken step of $0.1$.
As the previous results, the 8-direction bank of filters achieved higher values of correct classification rate compared to the 4-direction bank of filters.
In this case, a correct classification rate of $98.60\%$ was obtained, which clearly shows the effectiveness of the proposed method on the dynamic texture recognition.

\begin{table}[!t]
	\centering
		\begin{tabular}{|c|c|c|c|c|}
%			\cline{1-7} \cline{10-16}
			\hline
			\hline
			Speed Combination & 4-directions & 8-directions \\
			\hline
			$[0.1 - 0.2]$ &	92.50(3.40) &	94.92(3.09) \\
			$[0.1 - 0.5]$ &	96.00(2.40) &	96.82(2.41) \\
			$[0.1 - 1.0]$ &	96.56(2.11) &	98.02(1.65) \\
			$[0.1 - 1.5]$ &	96.92(2.32) &	\textbf{98.60(1.60)} \\
			$[0.1 - 2.0]$ &	97.24(2.18) &	97.84(1.92) \\
			$[0.1 - 2.5]$ &	96.92(2.49) &	97.34(2.36) \\
			$[0.1 - 3.0]$ &	96.37(2.98) &	96.94(2.73) \\

			\hline
			%\cline{1-7} \cline{10-16}			
		\end{tabular}
	\caption{Correct classification rate and standard deviation for different combinations of speed and direction on the dyntex database.}
	\label{tab:dyntex}
\end{table}

\section{Conclusion}
In this paper, we proposed a new method for dynamic texture recognition based on spatiotemporal Gabor filters. 
First, it convolves a sequence of images to a bank of filters and then extracts energy statistic from each response.
Basically, the spatiotemporal Gabor filters are built using speed $v$ and direction $\theta$.
The bank of filter is composed by filters with different values of speed and direction.

Promising results considering different combinations of $v$ and $\theta$ were achieved on two important databases: dyntex and traffic database.
On the traffic database, our method achieved a correct classification rate of $91.50\%$ using combination of high speeds.
On the other hand, a correct classification rate of $98.60\%$ was obtained on the dyntex database using a combination of low speeds.

\subsubsection*{Acknowledgments.}
WNG was supported by FAPESP grants 2010/08614-0.
BBM was supported by CNPq.
OMB was supported by CNPq grants 306628/2007-4 and 484474/2007-3.

%-------------------------------------------------------------------------
%\bibliographystyle{ieee}
%\bibliography{ieee}

\end{document}